# Mimicking the Behaviour of Idiotypic AIS Robot Controllers Using Probabilistic Systems

Amanda M. Whitbrook, Uwe Aickelin and Jonathan M. Garibaldi
Intelligent Modelling & Analysis Research Group,
School of Computer Science,
University of Nottingham, Nottingham, UK

**ABSTRACT**

Previous work has shown that robot navigation systems that employ an architecture based upon the idiotypic network theory of the immune system have an advantage over control techniques that rely on reinforcement learning only. This is thought to be a result of intelligent behaviour selection on the part of the idiotypic robot. In this paper an attempt is made to imitate idiotypic dynamics by creating controllers that use reinforcement with a number of different probabilistic schemes to select robot behaviour. The aims are to show that the idiotypic system is not merely performing some kind of periodic random behaviour selection, and to try to gain further insight into the processes that govern the idiotypic mechanism. Trials are carried out using simulated Pioneer robots that undertake navigation exercises. Results show that a scheme that boosts the probability of selecting highly-ranked alternative behaviours to 50% during stall conditions comes closest to achieving the properties of the idiotypic system, but remains unable to match it in terms of all round performance.

## 1. INTRODUCTION

An artificial immune system (AIS) is a computational algorithm that attempts to mimic properties of the vertebrate immune system in order to solve complex problems. There are a number of different types including clonal selection-based algorithms [4], negative selection-based algorithms [5] and idiotypic networks [6]. The latter group is inspired by Jerne's network theory of the immune system [1], which asserts that suppression and stimulation between antibodies plays an important role in the immune response.

Within the domain of mobile robotics, the idiotypic network has remained a popular choice of AIS, with most researchers opting to implement Farmer *et al.*'s computational model [2] of Jerne's theory. This is largely because the network of stimulation and suppression between antibodies (analogous to behaviours in these systems) is thought to provide a means of achieving a decentralized behaviour-selection mechanism for the robot. Initial results, for example [7-13] are certainly very encouraging, but lack any sort of comparison with other systems to assert the idiotypic advantage. Furthermore, the complex system dynamics are poorly understood.

However, in [3] the performance of a reinforcement learning (RL)-based idiotypic system is compared with a system that uses RL only, and provides statistical evidence that the idiotypic system is able to complete a task faster and with fewer stalls than the RL scheme. The paper also attempts to analyze the performance of both control systems in order to explain the idiotypic advantage. It suggests that the RL-only system provides a strategy that is too greedy, always selecting the behaviour (antibody) that best matches the current environmental situation (antigen). In contrast, the idiotypic system is much more flexible, allowing behaviours that are not necessarily the best to flourish. In particular, the paper proposes that when the robot is stalled the idiotypic mechanism is able to increase the rate of antibody change autonomously so that alternative behaviours are used instead of those already tried.

Given these suggestions, it is possible to create robot controllers that attempt to mimic these properties using probabilistic behaviour selectors coupled with RL. Hence, the main aim of this paper is the construction and testing of such systems in order to facilitate further scrutiny of idiotypic dynamics. In a number of experiments, nine different probabilistic schemes compete with the idiotypic system in order to establish that the idiotypic mechanism is not merely performing the equivalent of random behaviour arbitration, but acting in a more intelligent way. Furthermore, the gradual use of more complex systems that apply probabilistic selection more intelligently should help to provide additional insight into the processes that govern idiotypic dynamics.

This rest of this paper is arranged as follows. Section 2 provides some background information about Jerne's idiotypic network theory, Farmer's computation model of it, and the variation of the model used in [3]. Section 3 describes the architectures of the idiotypic and probabilistic systems that are used in this research, and illustrates the environments and problems used for testing them. The experimental procedures adopted are reported in Section 4, and Section 5 discusses the results obtained. Section 6 concludes the paper.

## 2. BACKGROUND

In the vertebrate immune system antibodies plays a central role in eliminating antigens (for example bacteria and viruses) from the body. In order to achieve this, an antibody's combining site (paratope) must be able to bind to a region of the antigen called the epitope. According to the clonal selection theory of Burnet [14], antibodies with paratopes that possess a good degree of match to a given antigen epitope pattern proliferate within the system, i.e. are cloned (increase in concentration) and are kept in circulation. However, Jerne's idiotypic network theory [1] proposes that antibodies also possess a set of epitopes called idiotopes, and that these are the mechanism by which antibodies recognize each other. He suggests that antibody concentration levels are also influenced by inter-antibody activity, i.e. antibodies that are recognized by others are suppressed and reduce in concentration and those that do the recognizing are stimulated and increase in concentration.

Farmer et al. [2] go on to suggest that the dynamics of an idiotypic AIS system with $L$ antigens $[y_1, y_2,..., y_L]$ and $N$ antibodies $[x_1, x_2,..., x_L]$ can be modelled with the following equation:

$$\dot{C}(x_i) = b\left[\sum_{j=1}^{L} U_{ij} C(x_i) C(y_j) - k_1 \sum_{m=1}^{N} V_{im} C(x_i) C(x_m) + \sum_{p=1}^{N} W_{ip} C(x_i) C(x_p)\right] \quad (1)$$
$$- k_2 C(x_i),$$

where $C$ represents concentration and $b$, $k_1$ and $k_2$ are constants. In Eq. (1) the first sum in the square bracket models antibody stimulation due to antigens, the second sum models inter-antibody suppression and the last sum models inter-antibody stimulation. The match specificities for these three kinds of interaction are given by some functions $U$, $V$ and $W$ respectively. N. B. The term outside the brackets embodies the natural antibody death rate. Here, the equation models both background antibody communication (i.e. that between all antibodies) and also active antibody communication. The latter is the stimulation and suppression that takes place between the antigenic antibody $\alpha$ (that with the best match to the presenting antigen) and any other antibody that matches the presenting antigen (i.e. any competing antibody).

In [3] background communication is ignored for simplification and antigen concentrations are not needed since each environmental scenario (antigen) is ranked in order of importance and weighted accordingly, i.e. multiple antigens are allowed to present themselves but one is deemed dominant and given greater weighting. Furthermore, $N \times L$ matrices $P$ and $I$, which represent the antibody paratope and idiotope respectively, are used. The elements of $P$ are the current RL scores, which reflect the degree of match between each antibody and antigen, and the elements of $I$ are fixed disallowed antibody-antigen combinations. Antibody communication is hence simulated by comparing the paratope of $\alpha$ with the idiotope of the competing antibodies (i.e. those that have nonzero match to the set of presenting antigens) and vice versa. The model thus reduces to Eq. (2) below:

$$\dot{C}(x_i) = b\left[\sum_{j=1}^{L} U_{ij} - k_1 \sum_{m=1}^{L} V'_{im} C(x_i) C(x_\alpha) + \sum_{p=1}^{L} W'_{ip} C(x_i) C(x_\alpha)\right] \quad (2)$$
$$- k_2 C(x_i),$$

with matching functions $U$, $V'$ and $W'$ given by:

$$U_{ij} = P[x_i, y_j] G(x_i)_j, \quad (3)$$

$$V'_{im} = P[x_\alpha, y_m] I[x_i, y_m] H_i, \quad (4)$$

$$W'_{ip} = (1 - P[x_i, y_p]) I[x_\alpha, y_p] H_i. \quad (5)$$

In Eq. (3) $G(x_i)$ is an antigen array with value zero for non-presenting antigens, value two for the dominant antigen $y_d$ if $P[x_i, y_d] > 0$, value zero for the dominant antigen if $P[x_i, y_d] = 0$, and value 0.25 for all other presenting antigens. In Eq. (4) and Eq. (5) $H$ is an antibody array with value one for competing antibodies and value zero for noncompeting antibodies.

However, Eq. (2) cannot be evaluated as a whole since $\alpha$ must be found first, so it is broken down into constituent parts:

$$\alpha = \max_i \left[S_1(x_i) = \sum_{j=1}^{L} P[x_i, y_j] G(x_i)_j\right], \quad (6)$$

$$S_2(x_i) = \sum_{m=1}^{L} P[x_\alpha, y_m] I[x_i, y_m] H_i C(x_i) C(x_\alpha), \quad (7)$$

$$S_3(x_i) = \sum_{p=1}^{L} (1 - P[x_i, y_p]) I[x_\alpha, y_p] H_i C(x_i) C(x_\alpha). \quad (8)$$

First, antibody $\alpha$ is computed from Eq. (6), then the suppression and stimulation factors are calculated from Eqs. (7) and (8) respectively. Finally, the global match strength $S_g$ is determined from:

$$S_g(x_i) = S_1(x_i) - k_1(S_2(x_i)) + S_3(x_i). \quad (9)$$

The concentration of each antibody is hence given by substitution of Eq. (9) into Eq. (2) giving:

$$\dot{C}(x_i) = b(S_g(x_i)) - k_2 C(x_i), \quad (10)$$

which is transformed to:

$$C(x_i)_{t+1} = C(x_i)_t + b(S_g(x_i)_t) - k_2 C(x_i)_t, \quad (11)$$

upon discretization. The antibody $\beta$ selected for execution is that with the highest normalized concentration, given by:

$$\beta = \max_i \|C(x_i)_{t+1}\|. \quad (12)$$

In [3] experiments are performed that vary the values of the suppression-stimulation balancing constant $k_1$ and the rate constant $b$ using a simulated robot navigation exercise as a test bed. Results show that when $k_2$ is fixed at 0.05, the robot tends to perform best with $b$ set approximately between 40 and 160 and $k_1$ set between 0.575 and 0.650. In this region $\alpha \neq \beta$ (there is an idiotypic difference) approximately 20% of the time. The parameter $k_2$ governs how quickly the antibodies reach zero concentration. In other systems this might lead to their removal and replacement with alternatives, but this particular architecture uses a fixed number of antibodies that are never replaced, so $k_2$ is deliberately kept low.

It is worth noting that in [17] a slightly different idiotypic design is used that allows for only one presenting antigen per iteration, and more importantly employs a variable idiotope matrix with probabilistic components. This means that the idiotypic difference rate is much harder to predict for given values in the $\{b, k_1, k_2\}$ space, and suggests that the findings in [3] may be altogether dependent on the choice of the fixed disallowed antibody-antigen combinations in the idiotope matrix. For this reason the same combinations used in the idiotope matrix in [3] are used here, see Section 3.

## 3. TEST ENVIRONMENT AND SYSTEM ARCHITECTURE

Throughout this research simulated Pioneer 3 robots are used with Player's Stage 2.0.3 simulator [15]. The virtual robots possess eight sonar sensors at the rear and a laser sensor at the front that spans 180º. For convenience this 180º sector is subdivided into six 30º subsectors 1 to 6, with 1 and 2 representing the left, 3 and 4 corresponding to the centre, and 5 and 6 corresponding to the right of the robot. A frontal camera that can detect different coloured objects is also placed centrally, so that the robot is able to recognize cyan squares placed in the doorways. Its task is to use these as markers in order to navigate through the rooms in two different maze environments, $M_1$ and $M_2$ which are shown in Figures 1 and 2 respectively. Note that when a robot has passed a cyan marker its path back to the previous room is blocked off manually.

As stated earlier, environmental information is modelled with antigens and robot behaviours are modelled as antibodies that possess a fixed action component, and an idiotope and paratope element value for each antigen. For this purpose, eight antigens and sixteen antibodies are created as detailed in Tables 1 and 2 respectively and as in [3] where justification for choosing them is also given. Table 1 shows the priority ranking of the antigens, with 0 the lowest (least urgent) and 5 the highest (most urgent). Detection of the various antigens is governed by several sensor reading metrics which include the minimum and average laser readings $Z_{min}$ and $Z_{av}$, the average rear sonar reading $E_{av}$, and the position of the minimum laser reading $R_{min}$. The maximum laser reading $Z_{max}$ is also used by antibody 11, see Table 2.

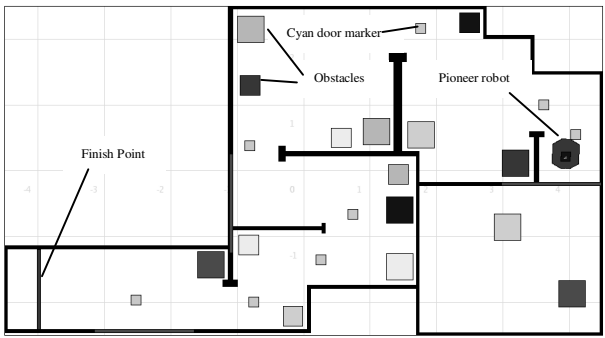

Figure 1. Simulated Maze World $M_1$

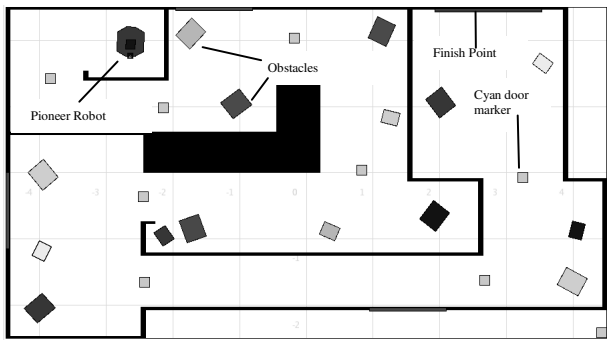

Figure 2. Simulated Maze World $M_2$

Ten control systems are created of which nine are probabilistic. The other system $I_D$ uses the idiotypic architecture described in Section 2, with $b$ set at 80, $k_1$ set at 0.65, and $k_2$ set at 0.05, as these parameter values fall within the region of $\{b, k_1, k_2\}$ space where performance is optimal in [3].

The $I_D$ controller begins by reading in the paratope matrix $P$ and the idiotope matrix $I$. Matrix $P$ is initially random, developing via RL as the code executes and the robot learns. Three different random paratope matrices are used $D_1$, $D_2$ and $D_3$, and each has initial element values between 0.50 and 0.75 to try to avoid any bias. Matrix $I$ is fixed and has value 0.50 for the antibody-antigen pairs 0-0, 0-2, 10-0, 10-2, 14-0, 14-2, 15-0 and 15-2, and value 1.00 for the pairs 1-2, 2-0, 3-1, 4-0, 5-2, 6-2, 7-0, 8-0, 9-2, 11-5, 12-5, 13-5, see Table 3. These pairs represent "disallowed" combinations, which effectively control the nature of the antibody connections within the idiotypic network.

Next the sensors are read and the dominant antigen and antigen array ($G$) element values are determined so that $S_1$ (the degree of match to antigen) can be calculated for each antibody and $\alpha$ can be determined using Eq. (6). Following this, Eqs. (7) and (8) are used to calculate suppression and stimulation respectively and thus deduce the global strength of match $S_g$ using Eq. (9). The concentration of every antibody in the system is then calculated using (11) and normalized using:

$$\|C(x_i)\|_{t+1} = \frac{C(x_i)_{t+1}}{\sum_{j=1}^{N} C(x_j)_{t+1}}, \quad (13)$$

so that the total number of antibody clones is kept constant to mimic the biology more closely and help prevent scaling problems. Note that the term concentration is used to mean the proportional number of clones of an antibody type in circulation if the total number of antibodies is held constant at $N$, with all concentration values beginning at 1.00. The final antibody selected is $\beta$ given by (12), i.e. $\beta$ is the antibody with the highest normalized concentration.

TABLE 1
THE ANTIGENS AND THEIR CONDITIONS

| Antigen (Priority) | Conditions |
|---|---|
| 0 - Object left (2) | $Z_{min} < 0.55$ m and $R_{min} = 1$ or 2 (-90º to -30º) |
| 1 - Object centre (2) | $Z_{min} < 0.55$ m and $R_{min} = 3$ or 4 (-29º to 29º) |
| 2 - Object right (2) | $Z_{min} < 0.55$ m and $R_{min} = 5$ or 6 (30º to 90º) |
| 3 - $Z_{av}$ > threshold (0) | $Z_{av} >= 0.45$ m |
| 4 - $Z_{av}$ < threshold (3) | $Z_{av} < 0.45$ m |
| 5 - Robot stalled (4) | Distance travelled = 0 |
| 6 - Robot blocked behind (5) | Distance travelled = 0 and $E_{av} < 0.35$ m |
| 7 - Door marker seen (1) | A cyan marker has been detected by the camera |

TABLE 2
THE ANTIBODIES AND THEIR ACTIONS

| | Antibody | Angular speed (º/s) | Linear speed (m/s) | Details |
|---|---|---|---|---|
| 0 | Reverse spin 1 | -90 | -0.15 | Fixed angular and linear movement |
| 1 | Slow right 15 | -15 | 0.06 | Fixed angular and linear movement |
| 2 | Slow left 15 | 15 | 0.06 | Fixed angular and linear movement |
| 3 | Fast centre | 0 | M/2 | Fixed linear movement |
| 4 | Fast left 15 | 15 | M/2 | Fixed angular and linear movement |
| 5 | Fast right 15 | -15 | M/2 | Fixed angular and linear movement |
| 6 | Slow right 35 | -35 | 0.06 | Fixed angular and linear movement |
| 7 | Slow left 35 | 35 | 0.06 | Fixed angular and linear movement |
| 8 | Fast left 35 | 35 | M/2 | Fixed angular and linear movement |
| 9 | Fast right 35 | -35 | M/2 | Fixed angular and linear movement |
| 10 | Reverse spin 2 | 90 | -0.15 | Fixed angular and linear movement |
| 11 | Wander max | Variable | M | Wander towards $Z_{max}$ reading |
| 12 | Wander min | Variable | M/2 | Wander away from $Z_{min}$ reading |
| 13 | Track blobs | Variable | M | Align towards centre of cyan marker |
| 14 | Reverse 1 | -25 | -0.15 | Fixed angular and linear movement |
| 15 | Reverse 2 | 25 | -0.15 | Fixed angular and linear movement |

$M$ = Maximum speed permitted (2.0 m/s)

When an antibody is selected for execution it carries out its designated action and the result of that action (half a second later) is scored either positively or negatively using RL. This means that paratope element value $P_{\beta d}$ is adjusted upon every iteration using:

$$(P_{\beta d})_{t+1} = \max[0, (P_{\beta d})_t + \tau_{t+0.5}], \qquad (14)$$

where $\tau$ is the positive or negative RL score awarded and $d$ is the index of the dominant antigen. Further details on the particular RL scheme used here are provided in [3], and a general explanation of RL can be found in [16].

The nine probabilistic systems, $R_1 - R_9$ are summarized in Table 4. They use the same essential architecture as the idiotypic system (described above), except that they compute antibody $\alpha$ only, not $\beta$, i.e. they omit the suppression and stimulation calculations in Eqs. (7) and (8). Having calculated $\alpha$ from the RL-scores of paratope matrix $P$ using Eq. (6), systems $R_1 - R_9$ either use it or simply select an alternative antibody $\mu$. The rate of $\mu$ selection and which alternative antibody is used both depend on pre-determined probability values. The idiotypic system is therefore mimicked by using a number of systems with probability values that simulate an approximate overall $\mu$ rate of 20%. Note that it is $P_{\alpha d}$ that is scored using RL for the probabilistic systems, or $P_{\mu d}$ when $\alpha$ is rejected in favour of alternative antibody $\mu$. Also, for systems $R_1 - R_9$ concentrations play no role in selecting the antibody that will execute its action.

In the case of $R_1$ there is a 20% chance of choosing any other antibody apart from $\alpha$, and these are selected with equal probability. System $R_2$ is similar with a 20% chance of not selecting $\alpha$, but the alternative antibody is chosen based on probabilities derived from the paratope matrix, i.e. the RL-scores representing the match between each antibody and the dominant antigen are used. The probability of selection $v$ of antibody $x_i$ is given by:

$$v(x_i) = \frac{P_{id}}{\sum_{j=1}^{N} P_{jd}}, \qquad (13)$$

where $N$ is the number of antibodies and $d$ is the index of the dominant antigen. If $\alpha$ is chosen again the process repeats until $\mu$ is different to $\alpha$.

Systems $R_3$, $R_4$ and $R_5$ also have a 20% $\mu$ rate, but when $\alpha$ is rejected $R_3$ always uses the antibody that is second-best and $R_4$ uses either the second or third best-matched antibody with equal probability. System $R_5$ uses either the second, third, or fourth best-matched antibody, but is twice as likely to use the second-best. System $R_6$ considers whether the previously-used antibody was deemed successful by the RL. If it was regarded as successful then there is only a 14% chance of selecting the second, third or fourth best-matched antibody. If it was marked as unsuccessful by the RL, then there is probably a greater need for a different antibody, so the probability of not selecting $\alpha$ increases to 28%. In either case, bias is toward choosing the second best-matched antibody, rather than the third or fourth.

Systems $R_7$, $R_8$ and $R_9$ are similar but take into account whether the robot is currently stalled or was stalled on the previous iteration. This methodology is adopted as previous analysis of antibody selection in system $I_D$ has shown that the idiotypic difference rate tends to increase to around 30% during stall conditions. With $R_7$, if there are no stall conditions then there is a 15% chance of not choosing $\alpha$. Again, bias is toward the second best-matched antibody, with the third and fourth best-matched being only half as likely to be selected. However, if the robot is currently stalled or was stalled on the previous iteration there is probably a much stronger requirement for an alternative antibody, so the chance of not selecting $\alpha$ increases to 33%, (bias is still toward the second-best antibody). Systems $R_8$ and $R_9$ work in the same way as $R_7$ but use 50% and 75% $\mu$ rates respectively when the robot is stalled and 13% and 2% $\mu$ rates otherwise. In systems $R_6$ to $R_9$ the probabilities are selected based on pre-trials to generate an approximate overall 20% observed $\mu$ rate.

## 4. EXPERIMENTAL PROCEDURES

Each of the ten control systems is run twelve times in Maze World $M_1$, six times starting with paratope $D_1$ and the other six times starting with paratope $D_2$. For each run, the time taken to complete the course $T$ is recorded along with the number of robot stalls $\sigma$. A stall represents a collision with an obstacle or the walls and is determined either by detecting that the robot has come to a complete stand-still for more than one time-loop interval (antigen 5) or by recording stand-still coupled with a rear-sonar reading of less than 0.35 m (antigen 6).

A fast robot that continually crashes or a careful robot that takes too long to complete the task is undesirable, so a fitness measure $F$, which combines $T$ and $\sigma$ is computed for each run. This is

TABLE 3
THE FIXED IDIOTOPE MATRIX

|    | 0    | 1    | 2    | 3    | 4    | 5    | 6    | 7    |
|----|------|------|------|------|------|------|------|------|
| 0  | 0.50 | 0.00 | 0.50 | 0.00 | 0.00 | 0.00 | 0.00 | 0.00 |
| 1  | 0.00 | 0.00 | 1.00 | 0.00 | 0.00 | 0.00 | 0.00 | 0.00 |
| 2  | 1.00 | 0.00 | 0.00 | 0.00 | 0.00 | 0.00 | 0.00 | 0.00 |
| 3  | 0.00 | 1.00 | 0.00 | 0.00 | 0.00 | 0.00 | 0.00 | 0.00 |
| 4  | 1.00 | 0.00 | 0.00 | 0.00 | 0.00 | 0.00 | 0.00 | 0.00 |
| 5  | 0.00 | 0.00 | 1.00 | 0.00 | 0.00 | 0.00 | 0.00 | 0.00 |
| 6  | 0.00 | 0.00 | 1.00 | 0.00 | 0.00 | 0.00 | 0.00 | 0.00 |
| 7  | 1.00 | 0.00 | 0.00 | 0.00 | 0.00 | 0.00 | 0.00 | 0.00 |
| 8  | 1.00 | 0.00 | 0.00 | 0.00 | 0.00 | 0.00 | 0.00 | 0.00 |
| 9  | 0.00 | 0.00 | 1.00 | 0.00 | 0.00 | 0.00 | 0.00 | 0.00 |
| 10 | 0.50 | 0.00 | 0.50 | 0.00 | 0.00 | 0.00 | 0.00 | 0.00 |
| 11 | 0.00 | 0.00 | 0.00 | 0.00 | 0.00 | 1.00 | 0.00 | 0.00 |
| 12 | 0.00 | 0.00 | 0.00 | 0.00 | 0.00 | 1.00 | 0.00 | 0.00 |
| 13 | 0.00 | 0.00 | 0.00 | 0.00 | 0.00 | 1.00 | 0.00 | 0.00 |
| 14 | 0.50 | 0.00 | 0.50 | 0.00 | 0.00 | 0.00 | 0.00 | 0.00 |
| 15 | 0.50 | 0.00 | 0.50 | 0.00 | 0.00 | 0.00 | 0.00 | 0.00 |

TABLE 4
THE PROBABILISTIC SYSTEMS

|       | Description | Probability of antibody selection (%) | | | | |
|-------|-------------|---|---|---|---|---|
|       |             | $\mu$ | $\alpha$ | 2nd best | 3rd best | 4th best |
| $R_1$ | 20% any other antibody but $\alpha$ - Equal probability | 20 | 80 | - | - | - |
| $R_2$ | 20% any other antibody but $\alpha$ - Probability based on RL score | 20 | 80 | - | - | - |
| $R_3$ | 20% 2nd best antibody | 20 | 80 | 20 | - | - |
| $R_4$ | 20% 2nd or 3rd best | 20 | 80 | 10 | 10 | - |
| $R_5$ | 20% 2nd, 3rd or 4th best | 20 | 80 | 10 | 5 | 5 |
| $R_6$ | If use of previous antibody deemed unsuccessful by RL | 28 | 72 | 14 | 7 | 7 |
|       | If use of previous antibody deemed successful by RL | 14 | 86 | 7 | 3.5 | 3.5 |
| $R_7$ | If current or previous antigen was either 5 or 6 | 33 | 67 | 16.5 | 8.25 | 8.25 |
|       | Otherwise | 15 | 85 | 7.5 | 3.75 | 3.75 |
| $R_8$ | If current or previous antigen was either 5 or 6 | 50 | 50 | 25 | 12.5 | 12.5 |
|       | Otherwise | 13 | 87 | 6.5 | 3.25 | 3.25 |
| $R_9$ | If current or previous antigen was either 5 or 6 | 75 | 25 | 37.5 | 18.75 | 18.75 |
|       | Otherwise | 2 | 98 | 1 | 0.5 | 0.5 |

given by:

$$F = \frac{1}{2}(T + \varphi_1 \sigma), \quad (14)$$

where $\varphi_1$ is the ratio of the mean task time to mean number of stalls over all the 120 experiments in $M_1$:

$$\varphi_1 = \frac{\overline{T}}{\overline{\sigma}}. \quad (15)$$

Maze World $M_2$ represents a more difficult task for the robot as there are more rooms with more obstacles and there is generally less space for the robot to move around in. The idiotypic system and the best-performing probabilistic controller from the experiments with $M_1$ (i.e. that with the best fitness) are both used for robot navigation in $M_2$, six times starting with $D_1$ and six times using $D_3$. Again, $T$ and $\sigma$ are recorded for each run and $F$ is calculated, this time using $\varphi_2$, the ratio of the mean task time to mean number of stalls over all the 24 experiments in $M_2$.

In both worlds, mean $T$, $\sigma$ and $F$ values are computed for each control system and are compared using a 1-tailed $t$-test, with differences accepted as significant at the 99% level only. As another measure of task performance, runs with an above average fitness for each world are counted as good and those with fitness in the bottom 10% of all runs in each world are counted as bad. In addition, the $\mu$ rate is noted for each run and the mean is calculated for each control system.

## 5. RESULTS

Table 5 shows the mean $T$, $\sigma$, $F$ and $\mu$ values for each control system in each world, and also the percentage of good and bad runs. Table 6 displays the significant difference levels when each of the systems is compared to the idiotypic controller.

TABLE 5
MEAN TASK TIME, COLLISIONS, FITNESS AND $\mu$ RATE
WITH PERCENTAGE OF GOOD AND BAD RUNS IN EACH SYSTEM

| System | Maze World | Means | | | | Run % | |
|---|---|---|---|---|---|---|---|
| | | $T$ | $\sigma$ | $F$ | $\mu$ (%) | Good | Bad |
| $I_D$ | | 218 | 21 | 180 | 21 | 92 | 0 |
| $R_1$ | | 414 | 62 | 419 | 20 | 25 | 25 |
| $R_2$ | | 317 | 39 | 293 | 19 | 50 | 0 |
| $R_3$ | | 295 | 55 | 335 | 19 | 42 | 17 |
| $R_4$ | $M_1$ | 290 | 45 | 298 | 18 | 58 | 17 |
| $R_5$ | | 296 | 43 | 296 | 19 | 58 | 8 |
| $R_6$ | | 313 | 54 | 342 | 21 | 42 | 17 |
| $R_7$ | | 302 | 42 | 296 | 19 | 58 | 8 |
| $R_8$ | | 259 | 39 | 263 | 20 | 67 | 8 |
| $R_9$ | | 293 | 48 | 312 | 16 | 50 | 8 |
| $I_D$ | $M_2$ | 339 | 18 | 258 | 17 | 83 | 0 |
| $R_8$ | | 398 | 46 | 479 | 21 | 25 | 33 |

The results show that none of the probabilistic controllers performs as well as $I_D$ in Maze World $M_1$. The idiotypic system has the fastest completion time, the least number of stalls and the best fitness. All of these performances are significantly better than the probabilistic systems, except in the case of $R_8$ and when comparing $\sigma$ values for system $R_7$ and $T$ values for $R_9$. Furthermore, $I_D$ has the highest percentage of good runs (92%) and has no bad runs. Probabilistic system $R_2$ also has no bad runs, but only 50% of its runs are considered good.

TABLE 6
SIGNIFICANT DIFFERENCE LEVELS BETWEEN THE PROBABILISTIC
SYSTEMS AND THE IDIOTYPIC SYSTEM

| System | Maze World | Significant Difference Level | | |
|---|---|---|---|---|
| | | $T$ | $\sigma$ | $F$ |
| $R_1$ | | 100.0 | 100.0 | 100.0 |
| $R_2$ | | 99.9 | 99.5 | 99.8 |
| $R_3$ | | 99.6 | 99.9 | 99.9 |
| $R_4$ | | 99.1 | 99.2 | 99.3 |
| $R_5$ | $M_1$ | 99.1 | 99.3 | 99.4 |
| $R_6$ | | 99.6 | 99.9 | 99.9 |
| $R_7$ | | 99.2 | 98.9 | 99.2 |
| $R_8$ | | 94.0 | 98.5 | 98.1 |
| $R_9$ | | 96.8 | 99.8 | 99.6 |
| $R_8$ | $M_2$ | 97.6 | 100.0 | 100.0 |

Since system $R_8$ is second best in terms of fitness, it is used in Maze World $M_2$ for comparison with $I_D$. However, in this world both its $\sigma$ and $F$ values have significantly higher means than those observed with $I_D$, and $T$ is almost significant. In addition, 33% of runs are deemed bad and only 25% are deemed good for system $R_8$. In contrast, 83% of the idiotypic system's runs are judged as good and none are judged as bad.

All of the probabilistic systems show an overall mean $\mu$ rate of approximately 20%, which validates the probability choices.

**Discussion**

The results achieved provide strong empirical evidence that the idiotypic system possesses a highly intelligent form of behaviour selection that cannot easily be mimicked using simple probabilistic systems. In fact, $I_D$ performs better even when a probabilistic system uses some form of inherent intelligence, for example basing the likelihood of antibody selection upon the current RL scores (as in $R_2$) or boosting the probability of selecting an alternative to $\alpha$ under certain conditions ($R_6 – R_9$).

The probabilistic controller that performs best in world $M_1$ (and therefore comes closest to mimicking idiotypic dynamics) is system $R_8$, which increases the theoretical $\mu$ rate (probability of selecting either the second, third, or fourth best-matched antibody) to 50% under stall conditions. However, the mean number of stalls is still significantly higher than for the idiotypic system in world $M_2$, which suggests that $R_8$ is less able to deal with more complex environments.

System $R_7$ has a theoretical $\mu$ rate of 33% during stall conditions, which is very close to the mean idiotypic difference rate recorded for $I_D$ under these circumstances (31%). However, its performance is inferior to $I_D$ and also to $R_8$, which increases the $\mu$ rate to 50% under stall conditions. This suggests that the idiotypic dynamics are doing more than merely raising the rate of antibody change when the robot is in difficulty. Indeed, [3] proposes that it is the increased RL success rate of the antibodies chosen during stall conditions that contributes to an idiotypic robot's superior performance, and that the idiotypic process works by selecting antibodies of similar type to $\alpha$. In other words, as well as raising the $\mu$ rate during stall conditions, the probabilistic systems also need a better mechanism for determining which alternative antibody should be selected. Presently, only the current second, third, and fourth best-matched antibodies are considered, with the second-best being twice as likely to be selected as the third or fourth. This is a fundamental weakness in the probabilistic schemes, as an alternative antibody with a highly-ranked RL score for a particular antigen does not necessarily represent an antibody with similar properties to $\alpha$. Further research is therefore needed, in particular, a detailed examination of the alternative antibodies that are chosen under stall conditions in the idiotypic system, and how they rank in terms of matching to the presenting antigens. If a general pattern of selection could be identified and formalized into a probabilistic algorithm that approximates it, it might be possible to mimic the idiotypic dynamics much more closely.

However, it is still questionable whether such a system would be able to equal or better the performance of $I_D$. This is because the idiotypic mechanism is a dynamic process of continuous change, where the behaviour selected at a given time affects future selections, i.e. it represents a self-regulating system with feedback. In contrast, the probabilistic systems are only flexible in that they permit other antibodies to be chosen; in all other aspects they are inherently rigid. Furthermore, feedback in the idiotypic system is driven by the use of concentrations in the choice of alternative antibody as well as global strength of match to antigen, which means that it provides a kind of memory feature for past selection as well as considering current environmental information. In fact, it may be the balance between these two aspects that gives idiotypic robots their advantage. In Eq. (10) parameter $b$ governs the weighting given to the global strength of match $S_g$ when calculating new concentration values, and experiments that vary this parameter have shown that idiotypic robots show significantly better performance when $b$ is within a certain region [3].

A probabilistic scheme that aims to imitate the dynamics of an idiotypic system accurately would therefore need to:

1. Incorporate some form of memory feature analogous to antibody concentrations that enables the system to record past antibody use.
2. Utilize a mechanism that gives weighted consideration to both the memory and the strength-of-match to antigen when selecting alternative antibodies. This would introduce feedback into the system and provide a more dynamic selection process.
3. Mimic the ideal idiotypic difference rates, both during stall conditions and when the robot is free.
4. Imitate the patterns of alternative antibody selection inherent in the idiotypic dynamics during stall conditions, ideally by using a method that favours antibodies with similar properties to $\alpha$.

This research has currently addressed item 3) only, which might explain why $R_8$ came closest to reproducing the performance of $I_D$. System $R_8$'s theoretical $\mu$ rate under stall conditions is greater (50%) than the corresponding idiotypic difference rate of $I_D$ (30%). The greater chance of switching to an alternative antibody may have provided some form of compensation for lack of the other features, and may account for $R_8$'s superior performance to $R_7$. However, it should be noted that system $R_9$, which boosted the $\mu$ rate to 75% under stall conditions, was inferior in performance to both $R_7$ and $R_8$. This suggests that there may be an optimal $\mu$ rate under stall conditions for probabilistic systems that lack design specifications 1), 2) and 4). Future research will investigate this further by determining the optimal value, incorporating the missing design features, and examining any changes in the optimal value once these are in place.

## 6. CONCLUSIONS

This research has compared the performances of an idiotypic AIS robot control system with nine other control systems that select robot behaviour using probability functions. It has provided substantial empirical evidence that the idiotypic selection mechanism is superior to any of these systems, which suggests that the idiotypic dynamics are facilitating more intelligent behaviour selection. The probabilistic system that comes closest to approximating these dynamics is one that boosts the likelihood of non-$\alpha$ selection (i.e. increases the $\mu$ rate) during stall conditions, although its performance is still inferior to the idiotypic system. This supports the notion that idiotypic behaviour arbitration incorporates an innate ability to recognize and respond effectively to situations in which the robot is trapped.

Further research will aim to study the patterns of alternative antibody selection within the idiotypic system during stall conditions, in particular the strength-of-match rankings of antibodies chosen instead of $\alpha$. Study of these patterns might show how idiotypic systems are able to nominate more successful antibodies, and how the selection-mechanism is able to determine which ones have similar properties to $\alpha$. This might enable a more accurate probabilistic model of the idiotypic system to be created. Furthermore, a means of recording past antibody use is absent in the probabilistic systems constructed here, and may contribute to their inferior performance. Thus, an important aspect of future research will be the construction of a probabilistic algorithm that imitates this additional feedback feature. A detailed examination of the relationship between antibody concentrations, past use and time in the idiotypic system would greatly assist in this process.